\documentclass{article}
\def\arxivbuild{}
\ifdefined\arxivbuild\else\documentclass{article}\fi 

\usepackage{iclr2027_conference,times}
\ifdefined\arxivbuild\iclrfinalcopy\fi

\usepackage[utf8]{inputenc} 
\usepackage[T1]{fontenc}    
\usepackage{hyperref}       
\usepackage{url}            
\usepackage{booktabs}       
\usepackage{amsmath}
\usepackage{amsfonts}       
\usepackage{pifont}       
\newcommand{\cmark}{\ding{51}}
\newcommand{\xmark}{\ding{55}}
\usepackage{nicefrac}       
\usepackage{microtype}      
\usepackage{xcolor}         

\usepackage{graphicx} 
\usepackage{comment}
\usepackage{hyperref}
\usepackage[capitalise,nameinlink]{cleveref}
\usepackage{placeins} 

\title{UniCaMo: Unified Camera and Motion Control for Video Diffusion Models}

\author{%
  \textbf{Long Vu}\thanks{Equal contribution.}\quad
  \textbf{Tan Ngo}\footnotemark[1]\quad
  \textbf{Animesh Karnewar}\quad
  \textbf{Amir Habibian} \\[3pt]
  \textbf{Binh-Son Hua}\thanks{Binh-Son Hua is affiliated with Trinity College Dublin, Ireland. Work done under consultancy capacity.}\quad
  \textbf{Hung Bui}\quad
  \textbf{Minh Hoai Nguyen}\quad
  \textbf{Phong Nguyen-Ha} \\[6pt]
  Qualcomm AI Research\thanks{Qualcomm AI Research is an initiative of Qualcomm Technologies, Inc.} \\[3pt]
  \texttt{\{vlong, tanngo, karnewar, ahabibia,} \\
  \texttt{hson, hungbui, minhhoai, phongnh\}@qti.qualcomm.com}
}

\begin{document}

\def\mA{\mathcal{A}}
\def\mB{\mathcal{B}}
\def\mC{\mathcal{C}}
\def\mD{\mathcal{D}}
\def\mE{\mathcal{E}}
\def\mF{\mathcal{F}}
\def\mG{\mathcal{G}}
\def\mH{\mathcal{H}}
\def\mI{\mathcal{I}}
\def\mJ{\mathcal{J}}
\def\mK{\mathcal{K}}
\def\mL{\mathcal{L}}
\def\mM{\mathcal{M}}
\def\mN{\mathcal{N}}
\def\mO{\mathcal{O}}
\def\mP{\mathcal{P}}
\def\mQ{\mathcal{Q}}
\def\mR{\mathcal{R}}
\def\mS{\mathcal{S}}
\def\mT{\mathcal{T}}
\def\mU{\mathcal{U}}
\def\mV{\mathcal{V}}
\def\mW{\mathcal{W}}
\def\mX{\mathcal{X}}
\def\mY{\mathcal{Y}}
\def\mZ{\mathcal{Z}} 

\def\bbN{\mathbb{N}} 
\def\bbR{\mathbb{R}} 
\def\bbP{\mathbb{P}} 
\def\bbQ{\mathbb{Q}} 
\def\bbE{\mathbb{E}}

\def\1n{\mathbf{1}_n}
\def\0{\mathbf{0}}
\def\1{\mathbf{1}}

\def\A{{\bf A}}
\def\B{{\bf B}}
\def\C{{\bf C}}
\def\D{{\bf D}}
\def\E{{\bf E}}
\def\F{{\bf F}}
\def\G{{\bf G}}
\def\H{{\bf H}}
\def\I{{\bf I}}
\def\J{{\bf J}}
\def\K{{\bf K}}
\def\L{{\bf L}}
\def\M{{\bf M}}
\def\N{{\bf N}}
\def\O{{\bf O}}
\def\P{{\bf P}}
\def\Q{{\bf Q}}
\def\R{{\bf R}}
\def\S{{\bf S}}
\def\T{{\bf T}}
\def\U{{\bf U}}
\def\V{{\bf V}}
\def\W{{\bf W}}
\def\X{{\bf X}}
\def\Y{{\bf Y}}
\def\Z{{\bf Z}}

\def\a{{\bf a}}
\def\b{{\bf b}}
\def\c{{\bf c}}
\def\d{{\bf d}}
\def\e{{\bf e}}
\def\f{{\bf f}}
\def\g{{\bf g}}
\def\h{{\bf h}}
\def\i{{\bf i}}
\def\j{{\bf j}}
\def\k{{\bf k}}
\def\l{{\bf l}}
\def\m{{\bf m}}
\def\n{{\bf n}}
\def\o{{\bf o}}
\def\p{{\bf p}}
\def\q{{\bf q}}
\def\r{{\bf r}}
\def\s{{\bf s}}
\def\t{{\bf t}}
\def\u{{\bf u}}
\def\v{{\bf v}}
\def\w{{\bf w}}
\def\x{{\bf x}}
\def\y{{\bf y}}
\def\z{{\bf z}}

\def\balpha{\mbox{\boldmath{$\alpha$}}}
\def\bbeta{\mbox{\boldmath{$\beta$}}}
\def\bdelta{\mbox{\boldmath{$\delta$}}}
\def\bgamma{\mbox{\boldmath{$\gamma$}}}
\def\blambda{\mbox{\boldmath{$\lambda$}}}
\def\bsigma{\mbox{\boldmath{$\sigma$}}}
\def\btheta{\mbox{\boldmath{$\theta$}}}
\def\bomega{\mbox{\boldmath{$\omega$}}}
\def\bxi{\mbox{\boldmath{$\xi$}}}
\def\bnu{\mbox{\boldmath{$\nu$}}}                                  
\def\bphi{\mbox{\boldmath{$\phi$}}}
\def\bmu{\mbox{\boldmath{$\mu$}}}

\def\bDelta{\mbox{\boldmath{$\Delta$}}}
\def\bOmega{\mbox{\boldmath{$\Omega$}}}
\def\bPhi{\mbox{\boldmath{$\Phi$}}}
\def\bLambda{\mbox{\boldmath{$\Lambda$}}}
\def\bSigma{\mbox{\boldmath{$\Sigma$}}}
\def\bGamma{\mbox{\boldmath{$\Gamma$}}}
                                  
\newcommand{\myprob}[1]{\mathop{\mathbb{P}}_{#1}}

\newcommand{\myexp}[1]{\mathop{\mathbb{E}}_{#1}}

\newcommand{\mydelta}[1]{1_{#1}}

\newcommand{\myminimum}[1]{\mathop{\textrm{minimum}}_{#1}}
\newcommand{\mymaximum}[1]{\mathop{\textrm{maximum}}_{#1}}    
\newcommand{\mymin}[1]{\mathop{\textrm{minimize}}_{#1}}
\newcommand{\mymax}[1]{\mathop{\textrm{maximize}}_{#1}}
\newcommand{\mymins}[1]{\mathop{\textrm{min.}}_{#1}}
\newcommand{\mymaxs}[1]{\mathop{\textrm{max.}}_{#1}}  
\newcommand{\myargmin}[1]{\mathop{\textrm{argmin}}_{#1}} 
\newcommand{\myargmax}[1]{\mathop{\textrm{argmax}}_{#1}} 
\newcommand{\myst}{\textrm{s.t. }}

\newcommand{\denselist}{\itemsep -1pt}
\newcommand{\sparselist}{\itemsep 1pt}

\definecolor{pink}{rgb}{0.9,0.5,0.5}
\definecolor{purple}{rgb}{0.5, 0.4, 0.8}   
\definecolor{gray}{rgb}{0.3, 0.3, 0.3}
\definecolor{mygreen}{rgb}{0.2, 0.6, 0.2}

\newcommand{\cyan}[1]{\textcolor{cyan}{#1}}
\newcommand{\red}[1]{\textcolor{red}{#1}}  
\newcommand{\blue}[1]{\textcolor{blue}{#1}}
\newcommand{\magenta}[1]{\textcolor{magenta}{#1}}
\newcommand{\pink}[1]{\textcolor{pink}{#1}}
\newcommand{\green}[1]{\textcolor{green}{#1}} 
\newcommand{\gray}[1]{\textcolor{gray}{#1}}    
\newcommand{\mygreen}[1]{\textcolor{mygreen}{#1}}    
\newcommand{\purple}[1]{\textcolor{purple}{#1}}       

\definecolor{greena}{rgb}{0.4, 0.5, 0.1}
\newcommand{\greena}[1]{\textcolor{greena}{#1}}

\definecolor{bluea}{rgb}{0, 0.4, 0.6}
\newcommand{\bluea}[1]{\textcolor{bluea}{#1}}
\definecolor{reda}{rgb}{0.6, 0.2, 0.1}
\newcommand{\reda}[1]{\textcolor{reda}{#1}}

\def\changemargin#1#2{\list{}{\rightmargin#2\leftmargin#1}\item[]}
\let\endchangemargin=\endlist
                                               
\newcommand{\cm}[1]{}

\newcommand{\mhoai}[1]{{\color{magenta}\textbf{[Hoai: #1]}}}

\newcommand{\mtodo}[1]{{\color{red}$\blacksquare$\textbf{[TODO: #1]}}}
\newcommand{\myheading}[1]{\vspace{0.5ex}\noindent \textbf{#1}}
\newcommand{\htimesw}[2]{\mbox{$#1$$\times$$#2$}}


%
%
%

\newcommand{\Sref}[1]{Sec.~\ref{#1}}
\newcommand{\Eref}[1]{Eq.~(\ref{#1})}
\newcommand{\Fref}[1]{Fig.~\ref{#1}}
\newcommand{\Tref}[1]{Table~\ref{#1}}

\maketitle
\ifdefined\arxivbuild\lhead{}\fi

\begin{abstract}
We present \textbf{UniCaMo}, an unified framework for joint control of camera and object motion in image-and-text-to-video diffusion models. Given a reference image, text prompt, sparse 3D object trajectories, and a 3D camera trajectory, UniCaMo constructs motion-consistent input noise that directly encodes the desired motion. Our 3D-grounded noise construction combines reference-derived noise propagation with a spherical Gaussian noise representation for newly revealed regions, providing a simple and effective motion prior without requiring reconstruction of unseen scene geometry. UniCaMo requires no architectural modifications or dedicated control modules and can be adapted using lightweight LoRA fine-tuning with low training data volume. Experiments on standard controllable video generation benchmarks demonstrate state-of-the-art motion controllability and high visual quality with minimal inference overhead. \ifdefined\arxivbuild\textbf{Qualitative videos are available on our project page: \url{https://phongnhhn.info/Unicamo/}.}\else\textbf{Qualitative videos are presented on the HTML webpage provided in the submitted supplementary zip files.}\fi

\end{abstract}

\section{Introduction}

Controllable video generation is important for applications ranging from visual content creation to robotics and embodied AI~\citep{wang2026robovip,jang2025dreamgen}, and has received increasing attention from the research community. Existing approaches generally start from a pretrained image-and-text-to-video diffusion model and finetune it to introduce controllability, broadly following two directions. One line of work adds adaptation or control modules that inject motion guidance into the generation process~\citep{chu2025wan,he2024cameractrl,bai2025recammaster,li2025magicmotion}. Motion constraints are typically injected late in the generation pipeline, when visual content has already begun to emerge, leaving limited flexibility to resolve motion ambiguity in a geometrically coherent manner while maintaining realism and temporal consistency. These methods also augment already large video diffusion models with additional architectural components, increasing model complexity and inference cost. An appealing alternative is to impose motion at the very beginning of generation by exploiting the input noise. Video diffusion models generate videos by progressively denoising an initial sequence of Gaussian noise maps; rather than modifying the model architecture, the desired motion can therefore be encoded directly into these input noise maps~\citep{burgert2025go,chang2025warped}.

\begin{figure}[t]
\centering
\includegraphics[width=1.0\linewidth]{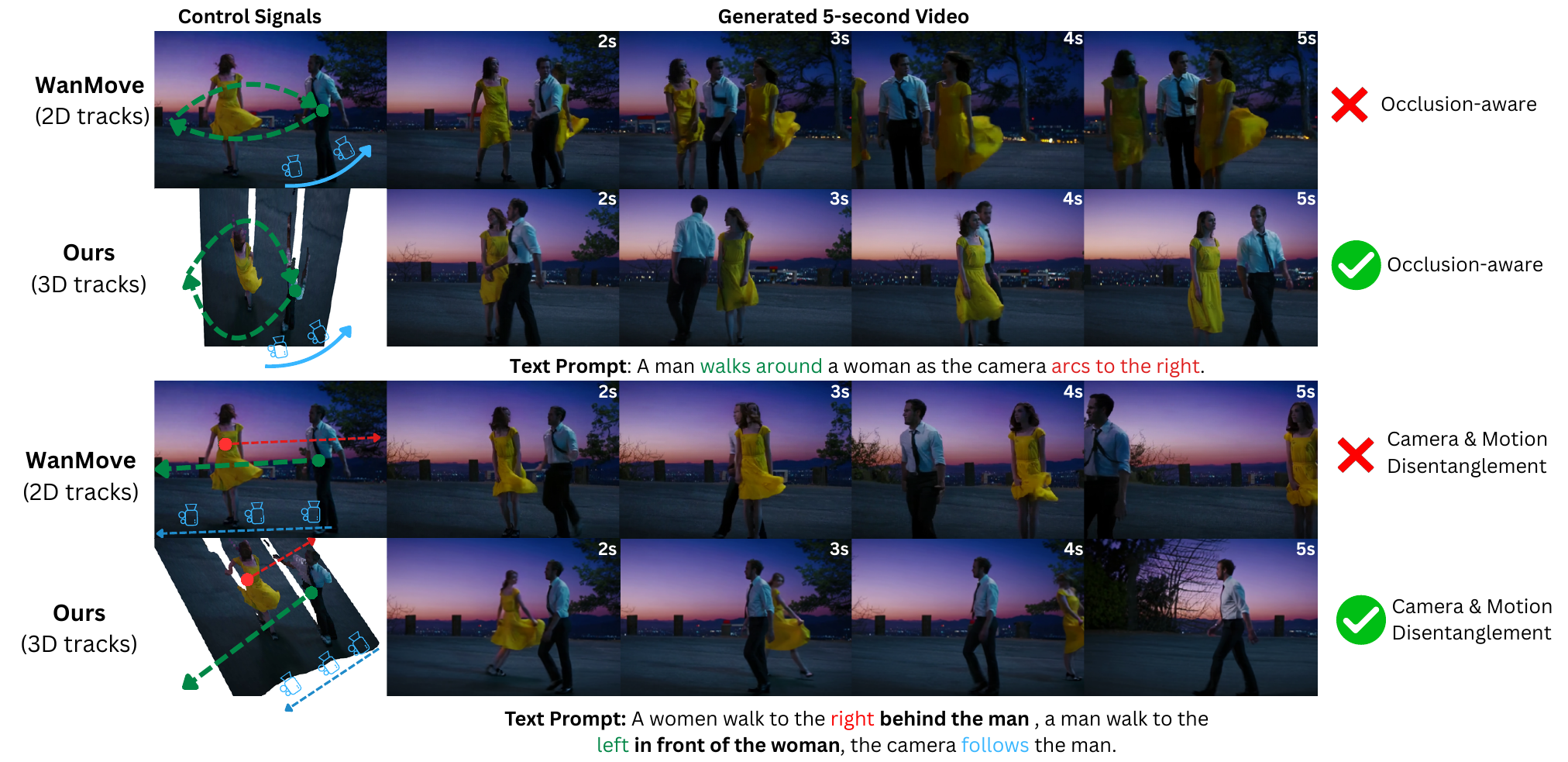}
    
    \caption{This teaser demonstrates a man walking and dancing around a woman while the camera pans to the right. UniCaMo jointly controls both object and camera motion through a 3D-grounded noise construction, enabling accurate occlusion handling and consistent appearance in unseen regions across frames, an effect that previous work with 2D cues~\citep{chu2025wan} cannot achieve.}
    \label{fig:teaser2}
\end{figure}

Existing input-noise-based methods, however, typically rely on 2D optical flow guidance and address object and camera motion separately. These limitations are closely related: the apparent motion of an object in a video is jointly determined by its 3D motion and the changing camera viewpoint, making it inherently ambiguous~\footnote{The HTML page provides a visualization of motion-noise between existing 2D and our 3D-aware approaches.} to specify and control the two independently in 2D image space. Moreover, complex motions—such as one person moving around another (Fig.~\ref{fig:teaser2})—are difficult to express without explicit depth and viewpoint information. These observations motivate the central question of this work: \emph{can we jointly control camera and object motions in 3D by manipulating only the input noise, without modifying the architecture of the underlying video diffusion model?}

To answer this question, we propose UniCaMo, a \underline{un}ified framework for joint \underline{ca}mera and object \underline{mo}tion control through view-and-motion-consistent input noise. UniCaMo lifts the reference image into 3D, where users specify the desired camera motion together with sparse 3D object trajectories. We construct the input noise according to these motions, using reference-derived noise where geometric correspondences are available and a spherical Gaussian noise representation to provide persistent noise for regions newly revealed by camera or object motion. Importantly, this construction need not recover the true geometry of unseen regions; it provides a plausible and geometrically consistent motion prior, while the pretrained video diffusion model supplies the visual and motion priors needed to synthesize realistic content. The resulting sequence of 2D noise maps directly initializes the video diffusion model, enabling joint 3D motion control without additional control modules or architectural modifications. Because the desired 3D motion is already explicitly encoded in the input noise, the model does not need to learn the motion-control mapping entirely from data, allowing effective adaptation with lightweight LoRA fine-tuning and a relatively small amount of computation for training.

Overall, UniCaMo offers three key advantages over existing controllable video generation approaches: \textbf{(1)} \emph{unified 3D control with lightweight adaptation}: UniCaMo jointly controls object and camera motion directly in 3D, avoiding the ambiguity of 2D motion specification and separate treatment of the two motions, requiring only lightweight LoRA adaptation; \textbf{(2)} \emph{simple and effective motion representation}: its 3D-grounded noise construction provides a plausible and geometrically consistent motion prior without requiring explicit reconstruction of unseen scene geometry; and \textbf{(3)} \emph{state-of-the-art performance}: UniCaMo achieves superior controllability and video quality over the evaluated baselines.

\section{Related Work}
\label{sec:related}

Video diffusion models have made remarkable progress since their early development based on 3D U-Net architectures~\citep{lumiere,svd,videocrafter1,animatediff}. Recent large-scale latent diffusion and transformer-based models~\citep{cogvideo,hunyuanvideo,wanvideo,cogvideox} achieve substantially improved visual fidelity, temporal coherence, and adherence to conditioning signals. As these models become increasingly capable, a natural next step is to provide finer control over the generated motion, particularly the motions of objects and the camera.

Existing motion-controlled video generation methods differ primarily in the motion representation and how it is incorporated into the diffusion model. For object motion, masks, boxes, and keypoints provide coarse spatial guidance, whereas optical flow and point trajectories enable finer-grained control~\citep{wang2024boximator,li2025magicmotion,geng2025motion,yin2023dragnuwa,zhang2025tora}. Camera motion is commonly represented by 6DoF poses or trajectories and incorporated through dedicated conditioning mechanisms~\citep{he2024cameractrl,wang2024motionctrl,bahmani2025ac3d}. More recent methods pursue joint object and camera control using trajectory-based conditioning, geometry-aware representations, or richer 3D information~\citep{li2025magicmotion,geng2025motion,chu2025wan,zheng2026versecrafter,zhang2026symphomotion}. These approaches generally introduce motion control through additional conditioning signals or modifications within the denoising process.

UniCaMo instead encodes object and camera motion directly into the input noise, connecting our work to structured noise initialization for controllable video generation. Noise-warping methods construct temporally correlated Gaussian noise by propagating noise according to motion fields. \citet{chang2025warped} highlight the importance of maintaining appropriate Gaussian noise statistics, while Go-with-the-Flow~\citep{burgert2025go} demonstrates efficient object motion control by warping input noise without an inference-time control module. Related work further shows that training with warped noise can induce useful equivariances and facilitate efficient sampling~\citep{liu2025equivdm}. These approaches, however, primarily operate with 2D motion fields or correspondences. UniCaMo extends this paradigm to joint 3D object and camera control by constructing view-and-motion-consistent noise: noise follows specified object trajectories while remaining consistent across changing viewpoints. A spherical noise representation provides persistent noise for newly revealed regions without explicitly reconstructing unseen scene geometry.

\section{UniCaMo -- Joint Camera and Object Motion Control}
\label{sec:method}

This section describes UniCaMo, our unified framework for joint camera and object motion control in video diffusion through view-and-motion-consistent input noise. UniCaMo requires no architectural modifications to the underlying video diffusion model, enabling lightweight fine-tuning with modest training data while introducing minimal additional overhead at inference time. We first provide an overview of the framework, followed by the details of noise construction and model training.

\subsection{Major components and overview}
\begin{figure}[t]
    \centering
  \includegraphics[width=1.0\textwidth]{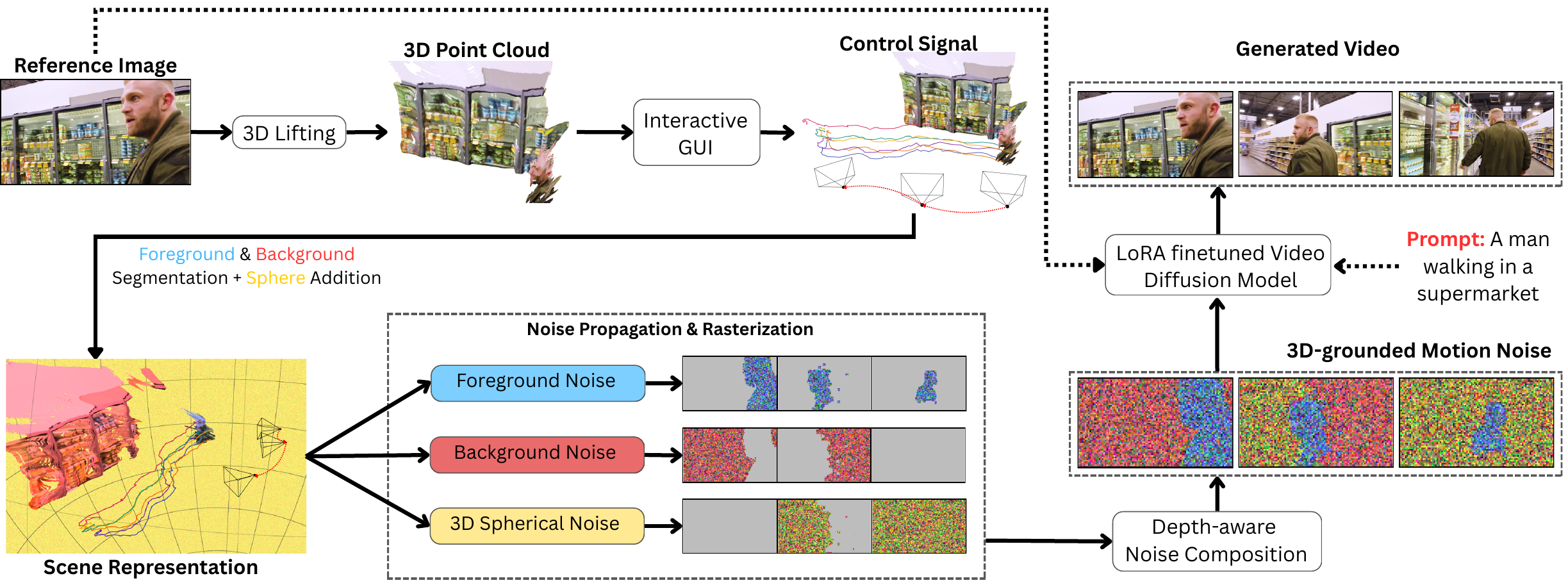}
  \caption{The UniCamo pipeline begins with 3D-lifting the input reference image to a 3D point cloud, within which per-frame control signals are defined via an interactive GUI, producing 3D point tracks and camera trajectories. From these, a 3D-grounded scene representation from three noise components: \textcolor{blue}{foreground noise} derived from 3D point tracks, \textcolor{red}{background noise} from depth cues, and \textcolor{yellow}{3D spherical noise} for persistent out-of-view anchors, which is then merged by a depth-aware noise composition step. The 3D-grounded noise is consumed by a LoRA finetuned video diffusion model to generate a video exhibiting corresponding object and camera motion. 
  }
  \label{fig:overall_pipeline}
\end{figure}

Our overall pipeline is shown in \cref{fig:overall_pipeline}, UniCaMo takes as input a reference image and a text prompt describing the desired scene and action, and outputs a generated video. In between, the framework comprises two stages: (1) an \emph{interactive interface} through which users specify the desired object and camera motions in 3D, and (2) a \emph{motion-guided generation model} that constructs motion-and-view-consistent noise from the reference image and specified motions to initialize an image-to-video diffusion model.

The \emph{interactive interface} first lifts the reference 2D image into a 3D point cloud, allowing the scene to be viewed and interacted with directly in 3D. Users specify object motion using sparse 3D point tracks on objects of interest, together with a desired 3D camera trajectory. This is an important distinction from prior approaches that specify motion using 2D point tracks in the 2D image plane, which are inherently ambiguous with respect to depth and viewpoint and can be insufficient for expressing complex motions, particularly when multiple objects cross or move around one another. By operating directly in 3D, UniCaMo provides a more expressive and geometrically explicit representation for jointly specifying object and camera motion.

The \emph{motion-guided generation model} builds upon a pretrained image-to-video diffusion model~\citep{wanvideo} that takes a sequence of Gaussian noise maps and progressively denoises them into video frames, conditioned on the reference image and text prompt. The input noise and output video therefore have a direct spatiotemporal correspondence: each noise map corresponds to an output frame, and spatial locations in the noise map influence corresponding regions of that frame. Prior work has shown that imposing coherent motion across the input noise maps can induce corresponding motion in the generated video \cite{burgert2025go, liu2025equivdm}. UniCaMo exploits this property by constructing \emph{motion-consistent noise maps} whose cross-frame correspondences reflect the user-specified 3D object and camera motions. Starting from a pretrained video diffusion model, we subsequently fine-tune it to better exploit these motion-consistent noise maps for controllable generation.

The key challenge is to construct consistent noise correspondences across frames given only the reference image and the user’s sparse 3D object trajectories and camera poses. Prior methods \citep{burgert2025go, liu2025equivdm} commonly rely on ground‑truth video to compute dense optical flow, which creates a systematic training–inference mismatch when such video is unavailable; our approach seeks to eliminate this discrepancy entirely. While noise associated with the reference image can be propagated according to the specified object and camera motions, these motions may reveal regions that are not visible in the reference frame and therefore have no corresponding noise values. UniCaMo addresses this problem by surrounding the reference point cloud with a \emph{3D spherical noise representation}, which provides persistent noise anchors for such unseen regions. Combining the reference-derived and spherical noise allows UniCaMo to construct a complete sequence of 2D noise maps consistent with the desired object and camera motions.

Importantly, the resulting noise sequence represents only \emph{one plausible realization} consistent with the desire motion. Recovering the true 3D geometry of occluded and out-of-view regions from a single reference image is fundamentally ill-posed and, more importantly, unnecessary for our purpose. Instead, the 3D sphere provides a simple and effective construction for assigning persistent noise values to otherwise undefined regions and maintaining their consistency across viewpoints. The sphere therefore does not represent the true scene geometry, nor does it constrain the generated scene to have spherical structure; it merely provides a geometrically consistent motion prior. The final video is determined jointly by this prior, the reference image, the text prompt, and the rich visual and motion statistics learned by the diffusion model from large-scale video data. Thus, rather than explicitly reconstructing unseen scene content, UniCaMo leverages the generative power of the pretrained model to synthesize plausible geometry and appearance while using the structured noise to guide the desired object and camera motions.

\subsection{Motion-Consistent Noise Construction}
\label{sec:noise_construction}

Given a reference image and the desired object and camera motions specified relative to it, our goal is to construct a sequence of 2D Gaussian noise maps whose cross-frame correspondences follow these motions. The motion specification consists of (i) a sparse set of target 3D point tracks $\mathcal{T}=\{\mathbf{x}_i^t\in\mathbb{R}^3\ | {i=\overline{1,N},t=\overline{1,T}}\}$ and (ii) a sequence of target camera poses $\mathcal{P}=\{\mathbf{P}^t | {t=\overline{1,T}}\}$, where $N$ is the number of point tracks and $T$ is the number of video frames. We first lift the reference image into a 3D point cloud using off-the-shelf 3D reconstruction methods~\cite{huang2025vipe,li2025megasam,wang2026vggtomega} and associate its points with Gaussian noise sampled for the reference frame.
Foreground noise is then propagated according to the specified object trajectories $\mathcal{T}$ and camera poses $\mathcal{P}$, while background noise is propagated according to $\mathcal{P}$ alone. This determines consistent noise values for regions represented by the reference point cloud, but leaves regions that are occluded or outside the reference field of view undefined. To complete these regions, we surround the reference point cloud with a 3D spherical noise representation that provides persistent noise values across viewpoints. Together, these steps produce a complete sequence of motion-consistent noise maps used to initialize the video diffusion model.

\noindent\textbf{Foreground noise propagation.}
We obtain a foreground mask of the reference image~\citep{carion2025sam3segmentconcepts} and retain the 3D point tracks in $\mathcal{T}$ associated with the foreground objects of interest. For each frame $t$, we project each 3D point $\mathbf{x}_i^t$ onto the image plane using the corresponding camera pose $\mathbf{P}^t$ and rescale the projected coordinates to the latent resolution. Specifically, given an image resolution of $W{\times}H$ and a latent resolution of $W_{\mathrm{lat}}{\times}H_{\mathrm{lat}}$, we scale the projected coordinates by $W_{\mathrm{lat}}/W$ and $H_{\mathrm{lat}}/H$. Let $\mathbf{u}_i^t\in\mathbb{R}^2$ denote the resulting latent-space location of $\mathbf{x}_i^t$. Starting from i.i.d.\ Gaussian noise sampled for the reference frame, we propagate noise along each trajectory by copying a $(2R-1){\times}(2R-1)$ patch centered at $\mathbf{u}_i^1$ in the first frame to its corresponding location $\mathbf{u}_i^t$ in each subsequent frame. We use $R=2$ by default, corresponding to a $3{\times}3$ patch. In this way, foreground noise follows both the specified 3D object motion and the changing camera viewpoint.

\noindent\textbf{Background noise propagation.}
For the background, we use the 3D point cloud obtained by unprojecting the reference image with its estimated depth and associate each point with an i.i.d.\ Gaussian noise value in place of its RGB color. The background point cloud remains fixed in world coordinates and is rasterized into each target frame according to the corresponding camera pose $\mathbf{P}^t$. Its projected noise therefore moves solely according to the camera trajectory. Thus, foreground noise follows both object and camera motion, whereas background noise follows camera motion alone.

\noindent\textbf{Spherical noise representation.}
The foreground and background propagation described above, however, cannot assign noise to regions newly revealed by object or camera motion, as these regions have no corresponding points in the reference point cloud. To provide temporally consistent noise for such regions, we surround the reference point cloud with a 3D sphere centered at the camera center of the reference frame. Let $\mathbf{C}^t\in\mathbb{R}^3$ denote the camera center at frame $t$, and let $d_{\max}^{0}$ denote the maximum estimated depth of the reference image. We set the sphere radius to $
r = \max\!\left( \tau \cdot \max_{t}\left\|\mathbf{C}^t-\mathbf{C}^1\right\|_2,\; d^{0}_{\max}\right)$, where $\tau=1.2$ provides a margin around the camera trajectory. We parameterize the sphere using latitude and longitude and assign its surface a fixed Gaussian noise texture $\mathbf{E}\in\mathbb{R}^{C\times H_{\mathrm{env}}\times W_{\mathrm{env}}}$, where $H_{\text{env}}$ and $W_{\text{env}}$ are proportional to the video resolution scaled by a factor $k=4$, and $C$ is the latent channel count of the video diffusion model.

For each frame, we cast a ray from the camera center through each latent location and compute its intersection with the sphere. The intersection point is converted to latitude–longitude coordinates, which are used to sample the corresponding value from the fixed Gaussian noise texture $\mathbf{E}$. Because all frames sample from the same spherical noise texture, rays intersecting the same spherical location receive the same noise value across viewpoints. The sphere therefore provides persistent and view-consistent noise for regions that cannot be associated with the reference point cloud. Details of the ray–sphere intersection are provided in the appendix.

\noindent\textbf{Depth-aware Noise Composition.}
We finally combine the foreground, background, and spherical noise contributions into a complete noise map for each frame. Multiple foreground and background points may project to the same latent location, particularly under object motion, occlusion, and viewpoint changes. Rather than resolving such collisions through arbitrary ordering~\citep{chu2025wan} or averaging~\citep{burgert2025go}, we exploit their explicit 3D geometry. For each latent location receiving multiple contributions from the foreground trajectories or background point cloud, we apply a z-buffer rule and retain the contribution with the smallest camera-space depth, i.e., the point closest to the camera. This enforces a geometrically meaningful visibility ordering and prevents noise associated with occluded surfaces from overwriting visible ones. Wherever this propagation provides a valid noise value, we retain it; locations that remain undefined are filled with the corresponding noise sampled from the sphere. This preserves the explicit geometric correspondences available from the reference image while using the spherical representation only where such correspondences are unavailable. The result is a complete sequence of motion noise maps consistent with both the specified 3D object trajectories and camera poses, providing the motion-aware initialization for subsequent video generation.

\section{Experiments}
\label{sec:experiments}

We evaluate UniCaMo in three settings: object motion control, camera motion control, and joint object and camera motion control. For object motion control, we use MoveBench~\citep{chu2025wan}, a recent benchmark comprising approximately 1,000 diverse, high-quality videos. For camera motion control, we randomly sample unseen 300 clips from RealEstate10K (RE10K)~\citep{zhou2018stereo}. Finally, to evaluate joint control, for which no suitable benchmark currently exists, we curate unseen 300 real-world videos from publicly available sources~\citep{Pexels_dataset} spanning diverse object motions and camera trajectories, following the setup in previous work~\citep{chu2025wan,zhang2026symphomotion}.

Using existing  protocol~\citep{chu2025wan,ren2025gen3c,wang2024motionctrl}, we evaluate three complementary aspects of the generated videos: \textbf{(i) visual quality}, measured by FID~\citep{fid}, FVD~\citep{fvd}, PSNR, and SSIM~\citep{ssim}; \textbf{(ii) object-motion fidelity}, measured by the endpoint error (EPE) of the generated trajectories~\citep{chu2025wan}; and \textbf{(iii) camera-motion fidelity}, measured by rotation error (RotErr), translation error (TransErr), and camera motion consistency (CamMC), which quantify the deviation between the generated and reference camera trajectories.

\subsection{Comparisons with State-of-the-Art Methods}

We compare UniCaMo against a broad range of prior motion-controllable video generation methods, covering object motion control, camera motion control, and joint control. Among these, methods such as Wan-Move~\citep{chu2025wan} and SymphoMotion~\citep{zhang2026symphomotion} support both object and camera motion control, while methods such as Go-with-the-Flow~\citep{burgert2025go} and ImageConductor~\citep{wang2024imageconductor} focus on object motion control, and CamI2V~\citep{zheng2024cami2v} and CameraCtrl~\citep{he2024cameractrl} focus on camera motion control. We generally evaluate these methods using their publicly released models. To enable more controlled comparisons, however, we additionally construct variants of relevant baselines that match UniCaMo in either the motion specification or the underlying video diffusion model. In particular, \textit{Wan-Move + TAPIP3D tracks} projects the same 3D object tracks used by UniCaMo onto the image plane and provides the resulting 2D tracks to Wan-Move, thereby matching the quality of the input trajectories. \textit{Go-with-the-Flow (Wan2.1-I2V-14B)} uses the same Wan2.1-I2V-14B backbone as UniCaMo and is retrained on the same data with the same training recipe, including the LoRA rank and training schedule, leaving the noise-construction strategy as the primary difference. These variants help disentangle the benefits of UniCaMo’s 3D motion specification from those of its motion-and-view-consistent noise construction.

\begin{figure}[t]
    \centering
  \includegraphics[width=1.0\textwidth]{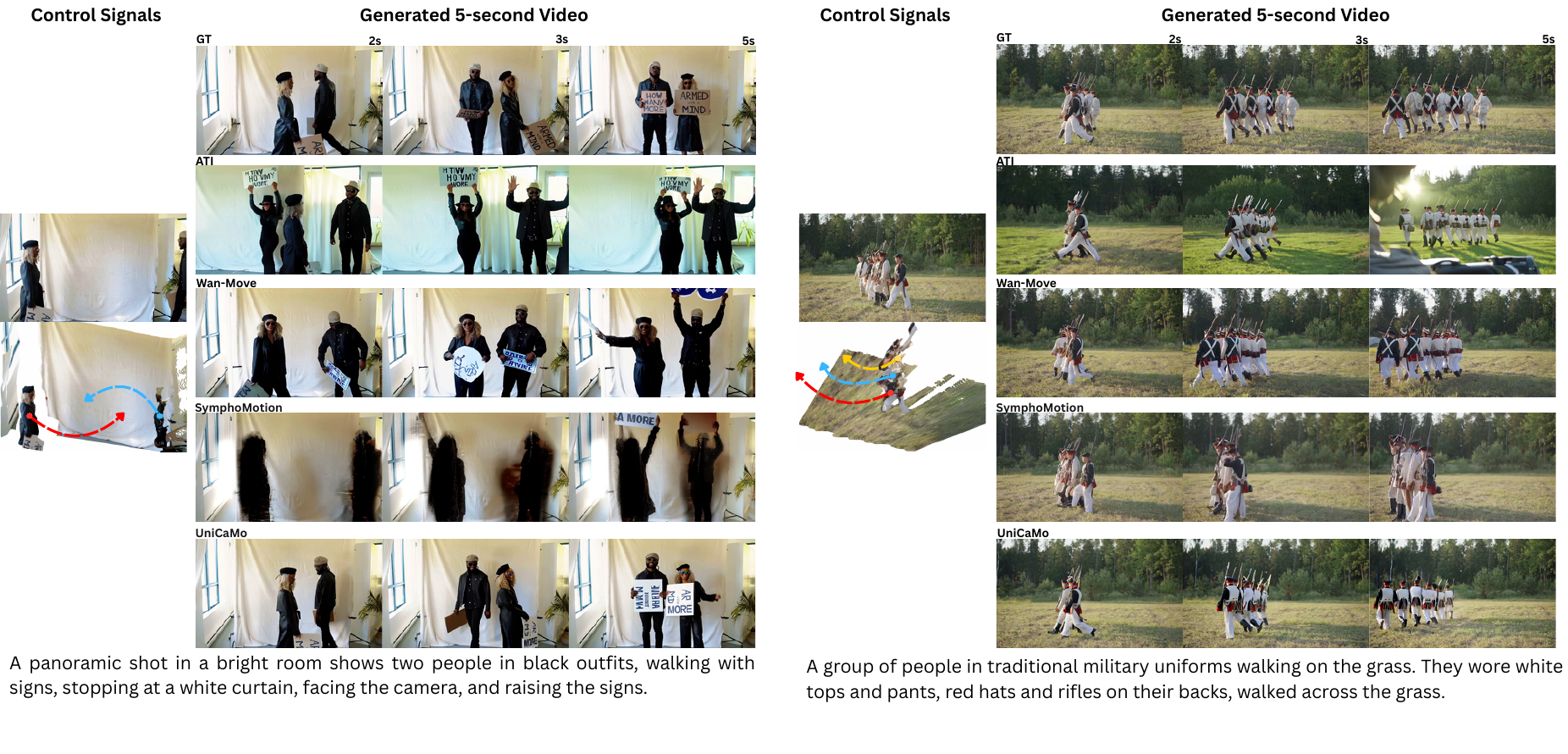}
  \caption{Qualitative results on MoveBench between UniCaMo and recent approaches~\citep{chu2025wan, wang2025ati, zhang2026symphomotion}. 
  }
  \label{fig:move-bench}
\end{figure}

\begin{figure}[t]
    \centering
  \includegraphics[width=1.0\textwidth]{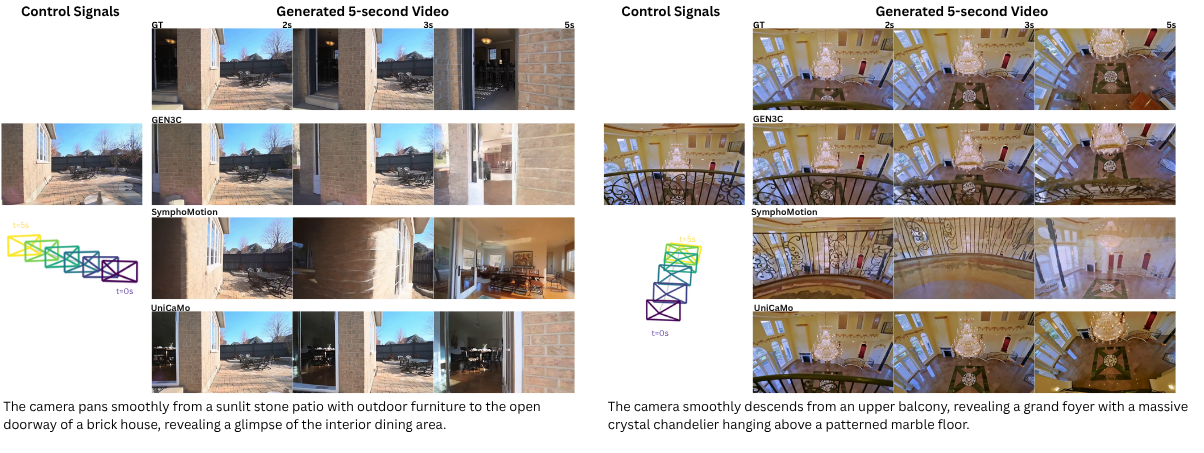}
  \caption{Qualitative results on RE10K between UniCaMo and other baselines ~\citep{zhang2026symphomotion, ren2025gen3c}. Our method enable more precise camera control than the baselines.}
  \label{fig:re10k_qual}
\end{figure}

\begin{figure}[t]
    \centering
  \includegraphics[width=1.0\textwidth]{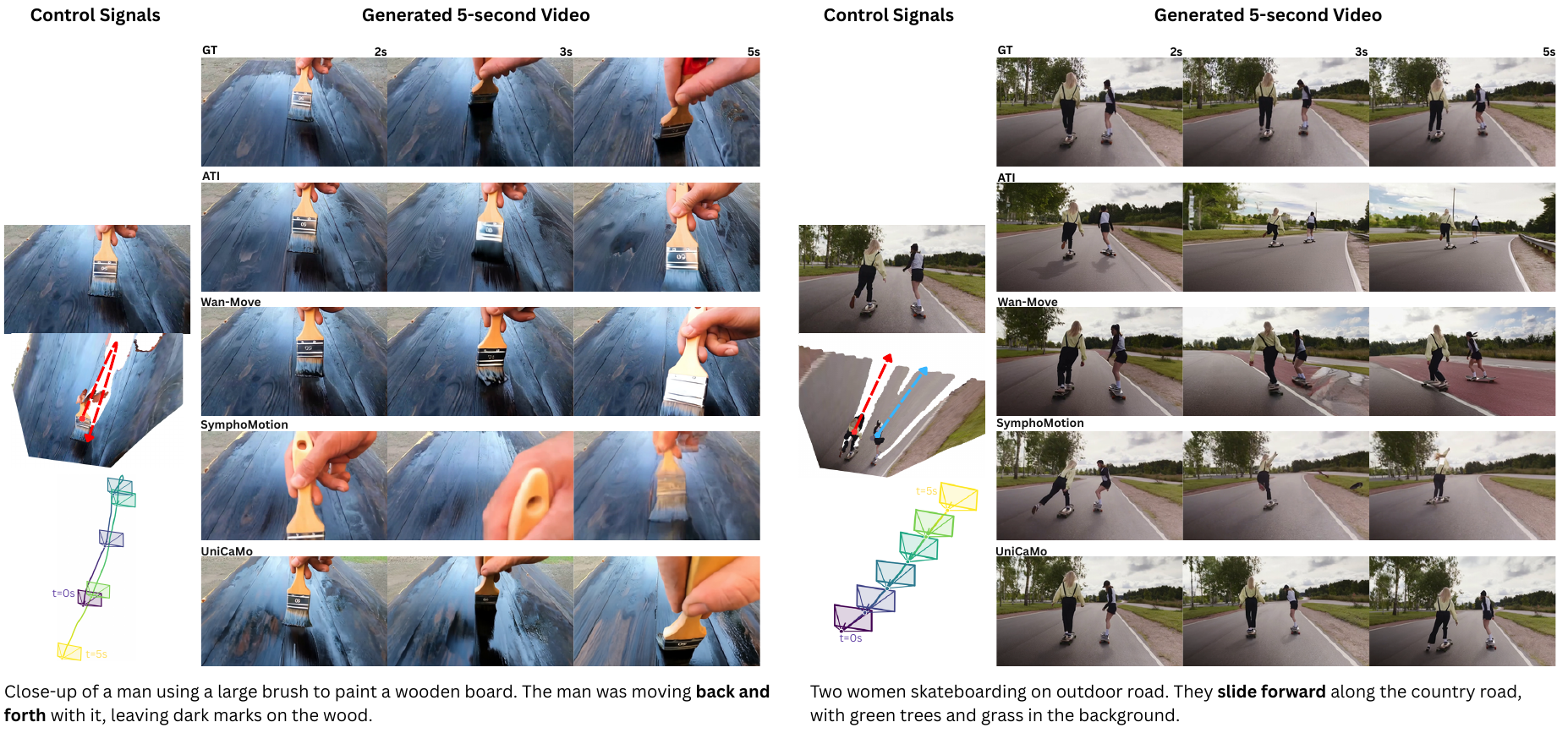}
  \caption{Jointly control over camera and object motions. While other works struggles to generate realistic object dynamics and accurate camera poses, UniCaMo achieves high-quality  control.}
  \label{fig:joint-bench}
\end{figure}

\begin{figure}[t]
    \centering
  \includegraphics[width=1.0\textwidth]{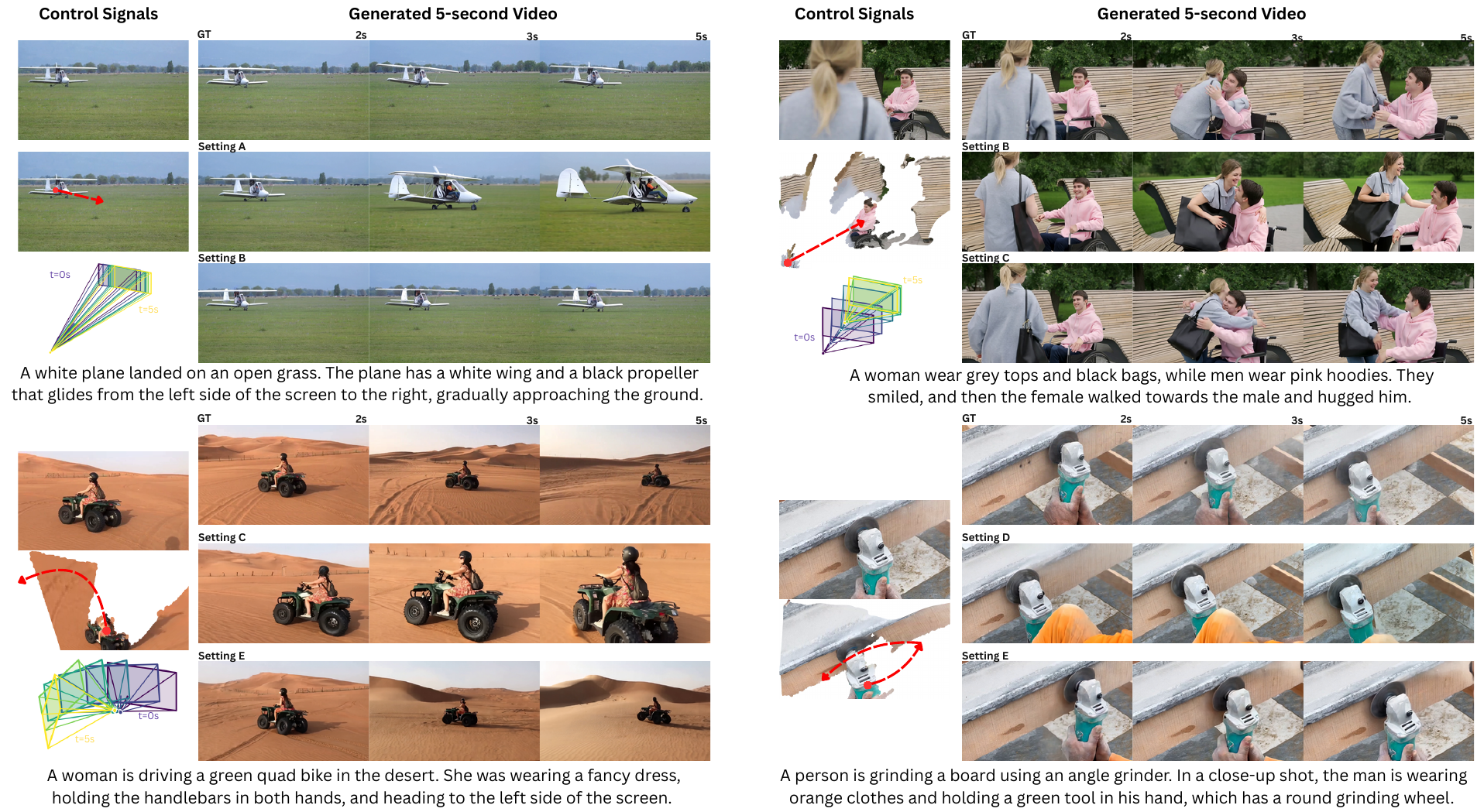}
  \caption{Qualitative results on ablation study of our 3D-aware structured motion noise.}
  \label{fig:abla}
\end{figure}


  


\begin{table}[t]
\centering
\caption{Evaluation results on MoveBench. We report quantitative comparisons on both the single-object and multi-object motion. Best results are shown in \textbf{bold}; second-best results are \underline{underlined}.}
\label{tab:movebench}
\setlength{\tabcolsep}{4pt}
\resizebox{\textwidth}{!}{%
\begin{tabular}{l|ccccc|ccccc}
\toprule
& \multicolumn{5}{c|}{\textit{Test cases with single-object motion}}
& \multicolumn{5}{c}{\textit{Test cases with multi-object motions}} \\
Methods
& FID $\downarrow$ & FVD $\downarrow$ & PSNR $\uparrow$ & SSIM $\uparrow$ & EPE $\downarrow$
& FID $\downarrow$ & FVD $\downarrow$ & PSNR $\uparrow$ & SSIM $\uparrow$ & EPE $\downarrow$ \\
\midrule
ImageConductor~\citep{wang2024imageconductor}
& 34.51 & 424.11 & 13.43 & 0.49 & 15.66
& 77.52  & 764.15 & 13.92 & 0.51 & 9.80 \\
Tora~\citep{zhang2025tora}
& 22.57 & 100.24 & 15.73 & 0.55 &  3.30
& 53.27  & 350.00 & 14.50 & 0.54 & 3.50 \\
ATI~\citep{wang2025ati}
& 11.27 & 105.35 & 15.44 & 0.58 & 5.09
& 30.72  & 278.91 & 16.19 & 0.62 & 3.17 \\
Go-with-the-Flow (CogX-5B)~\citep{burgert2025go}
& 12.49 & 216.90 & 15.80 & 0.62 &  3.04
& 35.72  & 399.40 & 16.93 & 0.60 & 3.71 \\
Go-with-the-Flow (Wan2.1-14B)
& 12.31 &  96.14 & 17.60 & 0.64 &  2.71
& 31.06 & 323.15 & 17.21 & 0.63 & 2.94 \\
Wan-Move~\citep{chu2025wan}
& 12.23 & \underline{83.50} & 17.80 & 0.64 &  2.60
& 28.80  & \underline{226.33} & 16.70 & 0.62 & 2.20 \\
Wan-Move + TAPIP3D tracks~\citep{chu2025wan}
& 12.11 &  87.10 & 18.10 & 0.67 & 2.54
& \underline{27.61}  & 294.80 & 16.50 & 0.65 & \underline{2.14} \\
SymphoMotion~\citep{zhang2026symphomotion}
& \textbf{9.92} & 89.27 & \textbf{18.50} & \textbf{0.74} & \underline{2.33} & 30.80 & 351.20 & \underline{17.57} & \underline{0.71} & 2.25 \\
UniCaMo (Ours)
& \underline{10.36} & \textbf{78.60} & \underline{18.40} & \underline{0.73} & \textbf{2.30}
& \textbf{26.12}  & \textbf{215.20} & \textbf{18.15} & \textbf{0.72} & \textbf{1.92} \\
\bottomrule
\end{tabular}%
}

\end{table}
\noindent\textbf{Object motion control.}
\Tref{tab:movebench} reports quantitative results on MoveBench, with results reported separately for single-object and multi-object motion. UniCaMo outperforms prior motion-control methods across all metrics, achieving lower FID and FVD, higher PSNR and SSIM, and lower trajectory error (EPE). The advantage is particularly pronounced on the 192 multi-object cases, where UniCaMo achieves the lowest FVD and EPE, demonstrating stronger adherence to the specified motions in more complex multi-object scenarios. Qualitative comparisons are shown in \Fref{fig:move-bench}.


\begin{table}[t]
\centering
\caption{
Camera-control evaluation on 300 randomly sampled clips from the RealEstate10K test set. We report rotation error (RotErr), translation error (TransErr), camera motion consistency (CamMC), and FID and FVD for visual quality. Best and second-best are \textbf{bolded} and \underline{underlined}.
}
\label{tab:camera}
\setlength{\tabcolsep}{6pt}
\scalebox{0.7915}{%
\begin{tabular}{lccccc}
\toprule
Methods & FID $\downarrow$ & FVD $\downarrow$ & RotErr $\downarrow$ & TransErr $\downarrow$ & CamMC $\downarrow$ \\
\midrule
MotionCtrl~\citep{wang2024motionctrl}
& 23.58 & 159.52 & 0.38 & 7.17 & 7.23 \\
CamI2V~\citep{zheng2024cami2v}
& 23.38 & 167.45 & \textbf{0.26} & 5.15 & 5.23 \\
CameraCtrl~\citep{he2024cameractrl}
& 23.80 & 165.37 & \underline{0.27} & 3.25 & 3.34 \\
Gen3C~\citep{ren2025gen3c}
& \underline{6.87} & \underline{52.39} & 0.33 & \underline{2.14} & \underline{2.28} \\
SymphoMotion~\citep{zhang2026symphomotion}
& 10.33 & 87.40 & 1.33 & 6.31 & 6.90 \\
UniCaMo (Ours)
& \textbf{5.99} & \textbf{39.93} & \textbf{0.26} & \textbf{1.89} & \textbf{1.98} \\
\bottomrule
\end{tabular}
}
\end{table}

\noindent\textbf{Camera control.}
Table~\ref{tab:camera} summarizes the camera-control results. UniCaMo achieves the lowest TransErr ($1.89$) and CamMC ($1.98$), while matching the best RotErr ($0.26$) and achieving FID $5.99$ and FVD $39.93$. These results support our design of motion-and-view-consistent noise: the spherical noise representation provides persistent noise correspondences across changing viewpoints, including newly revealed regions, enabling the generated video to better follow the specified camera trajectory. Qualitative comparisons are shown in \Fref{fig:re10k_qual}.

\begin{table}[t]
\centering
\caption{Evaluation of joint camera and object motion control on 300 clips randomly curated from publicly available sources, each containing simultaneous camera and object motion. Best results are shown in \textbf{bold}; second-best results are \underline{underlined}.}
\label{tab:joint}
\setlength{\tabcolsep}{6pt}
\resizebox{\textwidth}{!}{%
\begin{tabular}{lccccc|ccc}
\toprule
Method
& FID $\downarrow$ & FVD $\downarrow$ & PSNR $\uparrow$ & SSIM $\uparrow$ & EPE $\downarrow$
& RotErr $\downarrow$ & TransErr $\downarrow$ & CamMC $\downarrow$ \\
\midrule
ATI~\citep{wang2025ati}
& 27.83 & 336.25 & 14.58 & 0.54 & 8.64 & 18.59 & 46.41 & 57.81 \\
Wan-Move~\citep{chu2025wan}
& 23.81 & 294.49 & 15.28 & 0.60 & 5.03 & 8.84 & 44.41 & 48.49 \\
Wan-Move + TAPIP3D tracks~\citep{chu2025wan}
& 24.17 & 288.52 & 15.16 & 0.58 & 4.92 & 8.12 & 42.11 & 46.63 \\
SymphoMotion~\citep{zhang2026symphomotion}
& \underline{20.67} & \underline{219.86} & \underline{17.95} & \underline{0.70} & \underline{3.28} & \underline{2.34} & \underline{20.88} & \underline{21.96} \\
UniCaMo (Ours)
& \textbf{18.19} & \textbf{192.43} & \textbf{18.88} & \textbf{0.72} & \textbf{2.95} & \textbf{1.23} & \textbf{16.27} & \textbf{16.74} \\
\bottomrule
\end{tabular}%
}
\end{table}

\noindent\textbf{Joint camera and object motion control.}
We evaluate joint control on our 300 curated clips, each containing simultaneous object and camera motion, and report both object- and camera-control metrics in Table~\ref{tab:joint}. We use this curated dataset because MoveBench and RealEstate10K evaluate object and camera motion control separately and therefore do not capture the joint setting for which UniCaMo is designed. UniCaMo achieves the best performance across all eight metrics, with particularly strong gains in this more challenging setting, where the specified object trajectories and camera motion must be satisfied simultaneously. Qualitative comparisons are shown in \Fref{fig:joint-bench}.


\noindent\textbf{Supplementary videos.} Please check our provided HTML page and appendices to see more videos and comparisons.



\subsection{Ablation Studies}
\label{sec:ablation}

\begin{table}[t]
\centering
\caption{Ablation study of our noise construction method on 300 clips containing simultaneous camera and object motion. Best in \textbf{bold}; second-best \underline{underlined}.}
\label{tab:ablation_noise}
\setlength{\tabcolsep}{3pt}
\scalebox{0.7915}{%
\begin{tabular}{l|cccc|ccccc|ccc}
\toprule
Setting
& \shortstack{FG\\noise} & \shortstack{BG\\noise} & \shortstack{Sph.\\noise} & \shortstack{DA\\comp.}
& FID $\downarrow$ & FVD $\downarrow$ & PSNR $\uparrow$ & SSIM $\uparrow$ & EPE $\downarrow$
& RotErr $\downarrow$ & TransErr $\downarrow$ & CamMC $\downarrow$ \\
\midrule
A
& \xmark & \xmark & \xmark & \xmark & 21.59 & 238.41 & 15.58 & 0.62 & 8.23 & 14.45 & 43.78 & 52.86 \\
B
& \cmark & \xmark & \xmark & \cmark & 19.38 & 210.29 & 17.32 & 0.66 & 3.98 & 7.76 & 30.45 & 34.68 \\
C
& \cmark & \cmark & \xmark & \cmark & \underline{18.32} & \textbf{191.51} & 18.15 & \textbf{0.73} & \underline{3.16} & \underline{1.51} & \underline{18.42} & \underline{19.11} \\
D
& \cmark & \cmark & \cmark & \xmark & 18.51 & 196.23 & \underline{18.85} & 0.71 & 3.35 & 1.74 & 18.85 & 19.56 \\
E
& \cmark & \cmark & \cmark & \cmark & \textbf{18.19} & \underline{192.43} & \textbf{18.88} & \underline{0.72} & \textbf{2.95} & \textbf{1.23} & \textbf{16.27} & \textbf{16.74} \\
\bottomrule
\end{tabular}%
}
\end{table}

To assess the contribution of each component, we begin from purely random Gaussian noise (Setting A) and introduce our components one at a time, measuring the gain each provides over the preceding setting. Results are reported in \Cref{tab:ablation_noise}.

\noindent\textbf{Effect of foreground noise.}
Propagating noise along the foreground 3D tracks (B) show significant improvements over fully independent Gaussian noise (A). It more than halves the trajectory error and substantially reduces every camera metric, while also lowering FID and FVD. Motion written into the initialisation is therefore inherited by the generated video. The top-left of \Fref{fig:abla} shows the corresponding qualitative result: random noise still produces a coherent video rather than collapsing outright, but it no longer follows the trajectories specified by the user.

\noindent\textbf{Effect of background noise.}
Extending propagation from the foreground alone (B) to the background point cloud (C) benefits camera control. This is expected, since the background carries information about scene geometry and layout, and therefore a camera signal that foreground tracks alone cannot supply. The top-right of \Fref{fig:abla} illustrates this: Setting B still reproduces the correct object motion, but the camera fails to follow the user's intention.

\noindent\textbf{Effect of spherical noise.}
Adding the spherical representation for regions absent from the reference point cloud (E) yields a further camera-control gain over C, while FID and FVD remain essentially unchanged, indicating that the gain is specific to camera fidelity rather than to appearance. The bottom-left of \Fref{fig:abla} shows where this matters most: when the camera translates over a long distance or rotates through a large angle, extensive novel-view regions are revealed, and the spherical noise supplies most of their content.

\noindent\textbf{Effect of depth-aware noise composition.}
Resolving collisions with the depth-aware z-buffer rather than compositing in 2D (D) improves performance across the board: EPE drops from $3.35$ to $2.95$, RotErr from $1.74$ to $1.23$, TransErr from $18.85$ to $16.27$ and CamMC from $19.56$ to $16.74$, with FID and FVD also improving. Ordering contributions by camera depth therefore matters throughout the sequence, not only in explicitly occluded regions. The bottom-right of \Fref{fig:abla} shows the failure mode: without depth-aware composition, the motion noise of the grinder is overwritten and washed out by the background, leaving the object static in Setting D.

\section{Conclusions}
\label{sec:conclusion}
We presented UniCaMo, a unified framework for joint 3D camera and object motion control in image-to-video diffusion models. UniCaMo encodes the desired motion directly into the diffusion initialization by combining noise propagation along 3D object trajectories with a spherical noise representation that maintains persistent noise correspondences across changing viewpoints and newly revealed regions. This view-and-motion-consistent noise enables precise control without additional adaptation modules or architectural modifications to the underlying video diffusion model, requiring only lightweight fine-tuning. Experiments across object motion, camera motion, and joint control demonstrate that UniCaMo achieves strong controllability and visual quality, establishing structured noise initialization as a simple and effective approach to unified 3D motion control in video generation.
\FloatBarrier


\subsection*{AI use statement}
In this study, generative AI tools were used exclusively to correct grammatical and spelling errors. All AI-assisted edits were reviewed, and every modification proposed by these tools was independently verified by two authors. The authors accept full responsibility for the final content of the manuscript, including any text, claims, or artifacts produced with the assistance of generative AI.

\subsection*{Reproducibility statement}
We have provided key algorithms in the main text and the appendices. Full codebase, model checkpoints and datasets will be publicly available upon acceptance.

{
\small
\bibliographystyle{iclr2027_conference}
\bibliography{main}
}


\appendix

\clearpage
\section*{Technical appendices and supplementary material}


\section{User Interface}
\label{sec:UI}
\begin{figure}[h!]
    \centering
    \includegraphics[width=\linewidth]{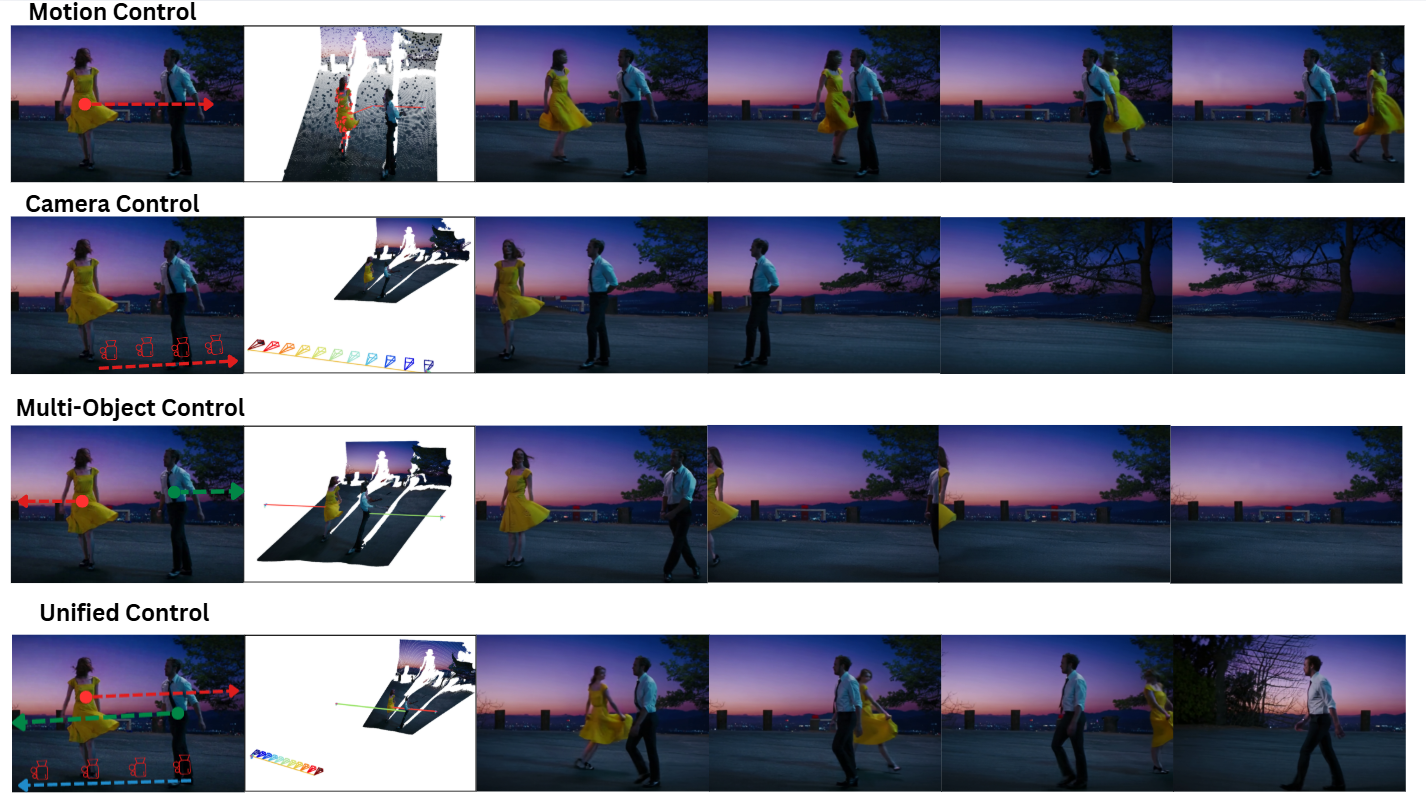}
    \caption{Visualization of our interactive GUI for unified camera and object motion control. Users select objects via polygon masks on the 2D image and edit 3D object trajectories and camera paths using interactive gizmos in the 3D view. Notice how 3D point-tracks allow us to define the exact 3D trajectory of the woman in yellow dress (go behind the man).}
    \label{fig:ui_demo}
\end{figure}

To support unified control of both object and camera motion at inference time, we implement an interactive graphical user interface (GUI) that enables users to define 3D object trajectories and 3D camera trajectories through an intuitive two-phase workflow. Fig.~\ref{fig:ui_demo} shows a visualization of our GUI and results.

\paragraph{Data Processing.}
Given a single input image, we first run ViPE~\citep{huang2025vipe} to estimate monocular depth and camera intrinsics. Using the estimated depth map and camera parameters, we reconstruct a dense 3D point cloud of the scene. This point cloud serves as the spatial context for all subsequent editing operations.

\paragraph{2D Object Selection.}
In the first phase, the GUI displays the input image as a 2D canvas. The user selects the object(s) of interest by drawing polygon masks directly on the image. Multiple objects can be selected as separate groups, each receiving its own mask. From each object mask, we proportionally sample 3D point tracks: for an object whose mask covers $n$ pixels out of the total $H \times W$ image, we allocate $\max(1, \lfloor n / (H \times W) \times N \rfloor)$ of the $N = 1024$ total point tracks. The remaining tracks are sampled uniformly from regions outside all masks to represent background motion. Each sampled pixel is mapped to its corresponding 3D world-space coordinate via the reconstructed point cloud.

\paragraph{3D Object Motion Editing.}
In the second phase, the GUI transitions to a 3D view showing the reconstructed point cloud alongside the sampled point tracks. The user defines target object trajectories by placing keyframes along a desired 3D path using interactive translation and rotation gizmos (6-DoF transform controls). Intermediate frames are obtained via linear interpolation between keyframes, producing continuous target motion trajectories for all points within each object group.

\paragraph{Camera Trajectory Editing.}
Camera motion is edited through a separate set of 6-DoF gizmos that control the virtual camera's position and orientation at user-specified keyframes. The camera trajectory is visualized as a sequence of frustums connected by a spline curve, allowing the user to intuitively design complex camera paths including panning, tilting, dollying, and orbiting. As with object motion, intermediate camera poses are linearly interpolated between keyframes.

\paragraph{Export and Video Generation.}
Once the user finalizes both object and camera trajectories, the GUI exports a single \texttt{.npz} file containing: (i) 3D point track coordinates of shape $(T, N, 3)$, (ii) per-frame visibility flags of shape $(T, N)$, (iii) camera intrinsics of shape $(T, 3, 3)$, and (iv) camera extrinsics (world-to-camera) of shape $(T, 4, 4)$. These exported signals are directly fed into our Guidance-based Noise Warping module (\Cref{sec:noise_construction}) to construct the 3D-grounded motion-consistent noise, which then initializes the video diffusion model for generation.

\section{Implementation Details}
\label{sec:supp_impl}




\subsection{Training and inference details}
\label{sec:details}
Given a ground-truth training video, we first preprocess it using off-the-shelf models to obtain the signals needed for motion-consistent noise construction. Specifically, we estimate camera intrinsics, extrinsics, and depth using ViPE~\citep{huang2025vipe}, foreground masks using SAM3~\citep{carion2025sam3segmentconcepts}, and sparse 3D point tracks using TAPIP3D~\citep{tapip3d}. These quantities capture the object and camera motions of the target video and allow us to apply the procedure in Sec.~\ref{sec:noise_construction} to construct a sequence of 2D Gaussian noise maps whose cross-frame correspondences are consistent with those motions. This naturally defines a supervised training pair: the motion-consistent noise sequence serves as the input initialization to the video diffusion model, while the original video serves as the target output. Importantly, the extracted geometric and motion signals are used only to construct the noise initializer and are not provided to the model as additional conditioning inputs.

We curate 400K training clips from DynPose100K++, WSDG-1M, and 4DNeX~\citep{huang2025vipe,chen20254dnex}, and precompute and cache all signals required for noise construction. To standardize captions across data sources, we re-caption all clips using Qwen2.5-VL~\citep{bai2025qwen3}, following the Wan-Move prompt-extension protocol~\citep{chu2025wan}. We fine-tune a pretrained Wan2.1-I2V-14B model~\citep{wanvideo} using flow matching, with the motion-consistent noise sequence as initialization and the reference frame and text prompt as conditioning. We apply LoRA only to the DiT denoiser, while freezing the VAE and all encoders. We use a LoRA rank of 64 with $\alpha{=}1$ and train for 10K steps in bfloat16 with gradient checkpointing on 32 NVIDIA H100 GPUs, using a batch size of 1 per GPU, for approximately three days. To improve robustness to different guidance densities, we randomly sample 512–1024 points from the extracted 3D tracks during training. At inference time, we use classifier-free guidance with $\gamma{=}5.0$. UniCaMo introduces no additional model parameters at inference time and incurs only approximately 0.2,s of noise-construction overhead per sample, preserving the runtime efficiency of the base model.

\subsection{Implementation Details of Foreground Noise Propagation}
\label{sec:track_the_noise_imple}
\paragraph{Patch copy--paste along trajectories.}
We initialize the first-frame latent noise as i.i.d. Gaussian,
\begin{align}
\mathbf{Z}_T^{1} \sim \mathcal{N}(\mathbf{0}, \mathbf{I}).
\end{align}
For each tracked point $i$ and frame $t$, let $(u_i^t, v_i^t)$ be its (integer) latent coordinates.
To propagate an identifiable noise footprint along the trajectory, we copy a small square patch around the track location in frame $1$ to the corresponding location in frame $t$.
Using a patch radius $r$, we define a patch of  $(2r{-}1)\times(2r{-}1)$ pixels where the copy follows:
\begin{align}
\mathbf{Z}_T^{t}\!\left[u_i^t+\Delta u,\; v_i^t+\Delta v\right]
:=
\mathbf{Z}_T^{1}\!\left[u_i^1+\Delta u,\; v_i^1+\Delta v\right].
\end{align}
where $\Delta u$ and $\Delta v$ are the pixel offsets. 
For $r=2$, the patch size is $3\times 3$, and the offset is
in $\{-1,0,1\}$. 
By transporting the same local noise pattern across time, motion becomes \emph{visible} to the latents so that visual features in the diffusion process are encouraged to follow the specified trajectory.
When multiple trajectories project to the same latent location at time $t$, naive composition yields ambiguous assignments and can introduce inconsistent motion cues. We then apply the Depth-aware Noise Composition rule described in \Cref{sec:noise_construction} to resolve the overlap.

\subsection{Implementation Details of Spherical Noise Sampling}
\label{sec:move_the_world_imple}
\paragraph{Ray construction (pixel $\rightarrow$ world ray).}
For each pixel $(p_x,p_y)$ (we use pixel centers), we form a unit ray direction in camera coordinates using intrinsics $(f_x,f_y,c_x,c_y)$:
\begin{align}
\mathbf{d}_{c}(p_x,p_y) 
&= 
\frac{1}{\left\| \left(\frac{p_x-c_x}{f_x},\,\frac{p_y-c_y}{f_y},\,1\right) \right\|_2}
\begin{bmatrix}
\frac{p_x-c_x}{f_x}\\[2pt]
\frac{p_y-c_y}{f_y}\\[2pt]
1
\end{bmatrix},
\end{align}
and then apply a camera-to-world transformation to determine the ray direction in the world space:
\begin{align}
\mathbf{d}_{w}(p_x,p_y) 
&= \mathbf{R}^t \, \mathbf{d}_{c}(p_x,p_y),
\end{align}
where $\mathbf{R}^t$ is the rotation from $\mathbf{T}_{c\rightarrow w}^t$.
The origin of the ray is at the camera location. 

\paragraph{Ray-sphere intersection and spherical UV.}
Let $\mathbf{S}$ be the sphere center with radius $r$. 
Let $\mathbf{o}^t$ be the ray origin and $\mathbf{d}_w$ be the ray direction in the world space. We define the direction from the sphere center $\mathbf{S}$ to the ray origin as $\mathbf{O}^t=\mathbf{o}^t-\mathbf{S}$.
We perform ray-sphere intersection by solving a quadratic equation.
With $b=\langle \mathbf{d}_w,\mathbf{O}^t\rangle$ and $c=\|\mathbf{O}^t\|_2^2-r^2$, the (non-negative) discriminant is
\begin{equation}
\Delta = b^2 - c,
\end{equation}
and we take the exiting intersection (consistent with the camera being inside or near the sphere):
\begin{equation}
\tau = -b + \sqrt{\Delta} .
\end{equation}
The hit point and outward normal are
\begin{equation}
\mathbf{P} = \mathbf{o}^t + \tau \mathbf{d}_w,
\qquad \mathrm{and} \qquad
\mathbf{n} = \frac{\mathbf{P}-\mathbf{S}}{r}.
\end{equation}
We convert the normal to spherical coordinates using longitude--latitude mapping:
\begin{align}
u &= \mathrm{fract}\!\left(\frac{\mathrm{atan2}(\mathbf{n}_x, \mathbf{n}_z)}{2\pi} + \frac{1}{2}\right),
\\
v &= \frac{\arccos(\mathrm{clip}(\mathbf{n}_y,-1,1))}{\pi},
\end{align}
where $\mathrm{fract}(\cdot)$ wraps $u$ into $[0,1)$.

\paragraph{Sampling sphere noise and downsampling to latent.}
We sample the Gaussian noise map at the computed spherical coordinates using nearest-neighbor lookup,
yielding a per-frame noise image.
This yields a 3D-consistent noise field while preserving the spatial Gaussianity per frame.

\section{Additional Ablation Studies}
\label{sec:additional_ablations}

\begin{table}[t]
\centering
\vspace{-0.2cm}
\caption{Analysis of \textit{(a)} the degradation rate $\theta$ for blending the constructed motion-consistent noise with Gaussian noise, \textit{(b)} the number of 3D point tracks, and \textit{(c)} the foreground noise patch dimension, evaluated on MoveBench evaluation data with multi-object motions.}
\label{tab:analysis_points_patch}
\renewcommand{\arraystretch}{1.15}
\setlength{\tabcolsep}{6pt}
\scalebox{0.7915}{%
\begin{tabular}{l|ccccc}
\toprule
\textbf{Setting}
& FID $\downarrow$ & FVD $\downarrow$ & PSNR $\uparrow$ & SSIM $\uparrow$ & EPE $\downarrow$ \\
\midrule
\multicolumn{6}{c}{\textit{(a) Degradation rate $\theta$}} \\
\midrule
$\theta = 0.0$ & 26.93 & 235.3 & \textbf{18.35} & \textbf{0.72} & \textbf{1.88} \\
$\theta = 0.2$ & 26.1 & 215.2 & 18.15 & \textbf{0.72} & 1.92 \\
$\theta = 0.5$ & \textbf{26.03} & \textbf{212.78} & 17.35 & 0.67 & 2.12 \\
\midrule
\multicolumn{6}{c}{\textit{(b) Number of 3D point tracks}} \\
\midrule
$N = 1024$ & \textbf{26.1} & \textbf{215.2} & \textbf{18.15} & \textbf{0.72} & \textbf{1.92} \\
$N = 512$  & 26.5 & 223.3 & 16.44 & 0.65 & 2.18 \\
$N = 256$  & 26.3 & 224.6 & 16.12 & 0.62 & 2.38 \\
\midrule
\multicolumn{6}{c}{\textit{(c) Patch dimension in Foreground Noise Propagation}} \\
\midrule
$R = 1.0$ & 27.6 & 254.6 & 15.32 & 0.54 & 3.34 \\
$R = 2.0$ & \textbf{26.1} & \textbf{215.2} & \textbf{18.15} & \textbf{0.72} & \textbf{1.92} \\
$R = 3.0$ & 26.8 & 221.19 & 16.70 & 0.64 & 2.28 \\
\bottomrule
\end{tabular}
}
\end{table}

\subsection{Effect of various hyperparameters.}
\label{sec:robust_3d}
\begin{figure}[!tbp]
    \centering
    \includegraphics[width=0.8\linewidth]{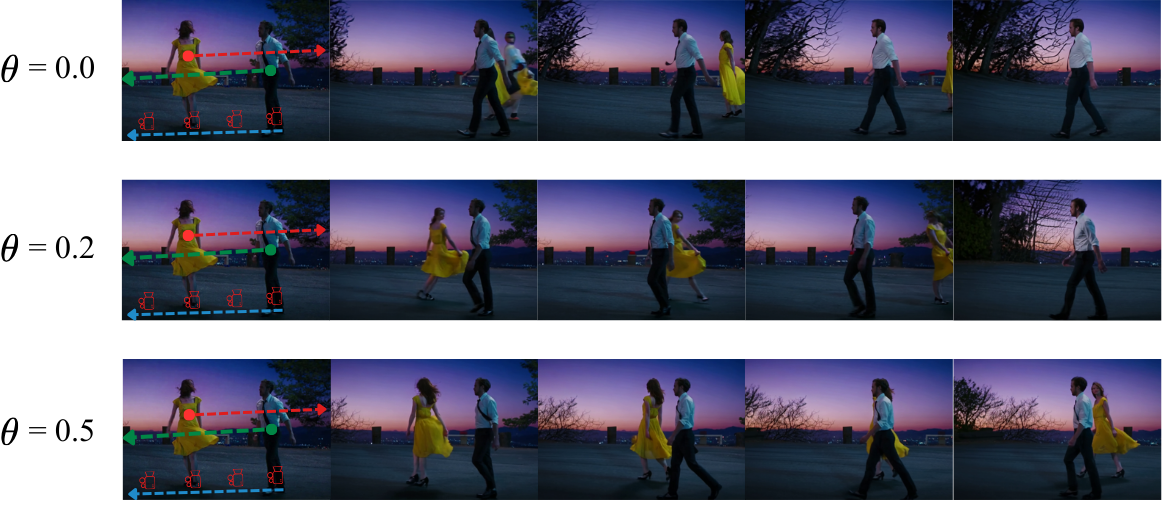}
    \caption{Effect of degradation rate on generated samples. A lower degradation rate ($\theta = 0.0$) offers more precise trajectory control but introduces visible artifacts. Conversely, a higher degradation rate ($\theta = 0.5$) yields visually plausible videos but reduces 3D trajectory accuracy. We adopt $\theta = 0.2$ as the default, which achieves the best trade-off between motion fidelity and visual quality.}
    \label{fig:degra_qual}
\end{figure}

\noindent\textbf{Degradation rate.}
Our motion-consistent noise is spatially Gaussian but introduces cross-frame correlations that differ from the independent Gaussian prior used to train the backbone. We therefore blend it with independent Gaussian noise using a randomly sampled degradation rate $\theta \in [0,0.5]$. At $\theta=0$, the initialization preserves the motion signal most strongly, yielding the best control fidelity (EPE $1.88$, PSNR $18.35$), but also deviates most from the original noise prior, resulting in lower video quality (FID $26.93$, FVD $235.3$). Increasing $\theta$ reduces this mismatch and monotonically improves FID and FVD, at the cost of motion-control fidelity. We adopt $\theta=0.2$ as a good trade-off between video quality and control fidelity (\Cref{tab:analysis_points_patch}~(a) and \Fref{fig:degra_qual}).

\begin{figure}[t]
    \centering
    \includegraphics[width=0.9\linewidth]{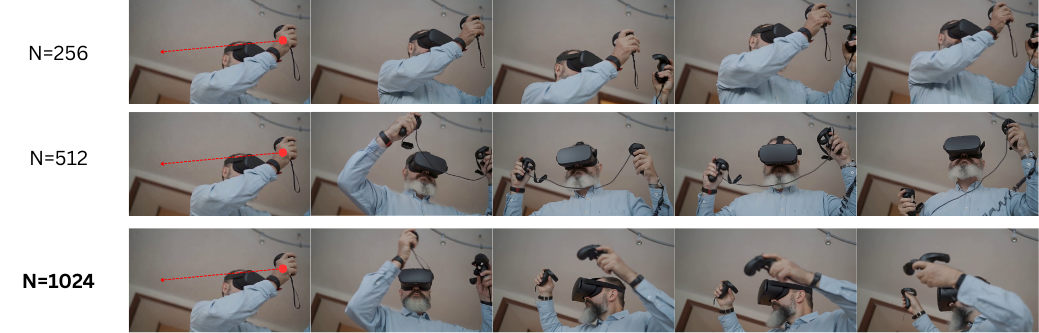}
    \caption{Qualitative results of our proposed UniCaMo method using different number of 3D point tracks.}
    \label{fig:supp_numtrack_qual}
\end{figure}

\noindent\textbf{Point tracks.}
We extract $1024$ tracks per video using TAPIP3D~\citep{tapip3d} and vary the number used during training to assess robustness to the track budget (\Cref{tab:analysis_points_patch}(b) and Fig.~\ref{fig:supp_numtrack_qual}). Visual quality is relatively insensitive to this choice, with FID and FVD varying by at most $0.4$ and $9.4$, respectively. Control fidelity is more sensitive: PSNR decreases from $18.15$ with $1024$ tracks to $16.44$ with $512$ and $16.12$ with $256$, while EPE increases from $1.92$ to $2.38$. We therefore use $1024$ tracks by default, although performance degrades gracefully as the number of tracks decreases.

\begin{figure}[t]
    \centering
    \includegraphics[width=0.9\linewidth]{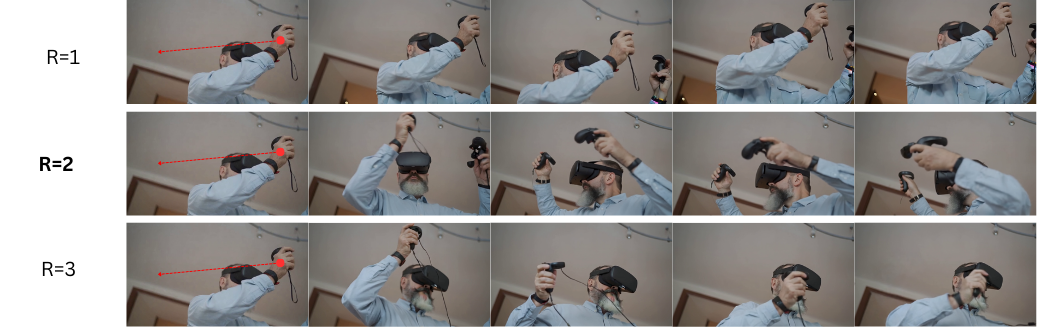}
    \caption{Qualitative results of our proposed UniCaMo method using different configuration of patch size $R$.}
    \label{fig:supp_radius_qual}
\end{figure}

\noindent\textbf{Patch size.}
We vary the patch dimension $(2R-1)$ used to propagate noise along the 3D point tracks; results are reported in \Cref{tab:analysis_points_patch}(c) and Fig.~\ref{fig:supp_radius_qual}. $R=2$ performs best across all five metrics. Both a smaller patch ($R=1$) and a larger one ($R=3$) degrade performance, suggesting that the patch should be large enough to provide a sufficiently strong motion signal while remaining sufficiently localized to avoid interference between neighboring trajectories. We therefore use $R=2$ by default.

\subsection{Robustness to Errors in 3D Estimation}
\label{sec:robust_3d}

UniCaMo depends on off-the-shelf modules for depth, camera pose and 3D point tracking, both when constructing training data and when preparing inference guidance, and errors in those estimates propagate into the constructed noise. To make the model tolerant of them, we add a small perturbation to the clean motion-visible noise during both training and inference. We assess that tolerance in two ways on the multi-object split of MoveBench (\Cref{tab:robust_3d}).

\begin{table}[t]
\centering
\caption{Robustness of UniCaMo to errors in the estimated 3D signals, evaluated on the multi-object split of MoveBench. \textit{(a)} perturbs the per-pixel depth by a controlled percentage; \textit{(b)} substitutes the camera poses and depth of different 3D foundation models. Best result in each block in \textbf{bold}.}
\label{tab:robust_3d}
\renewcommand{\arraystretch}{1.15}
\setlength{\tabcolsep}{6pt}
\scalebox{0.7915}{%
\begin{tabular}{l|ccccc}
\toprule
\textbf{Setting}
& FID $\downarrow$ & FVD $\downarrow$ & PSNR $\uparrow$ & SSIM $\uparrow$ & EPE $\downarrow$ \\
\midrule
\multicolumn{6}{c}{\textit{(a) Perturbation of the estimated depth}} \\
\midrule
Depth perturbation $= 0\%$ (default) & \textbf{26.08} & \textbf{215.20} & \textbf{18.15} & \textbf{0.72} & \textbf{1.92} \\
Depth perturbation $= 5\%$           & 26.12 & 215.21 & 18.11 & 0.71 & 1.94 \\
Depth perturbation $= 10\%$          & 27.14 & 223.56 & 17.35 & 0.68 & 2.10 \\
Depth perturbation $= 30\%$          & 31.20 & 283.32 & 16.22 & 0.60 & 2.57 \\
\midrule
\multicolumn{6}{c}{\textit{(b) Source of the estimated camera poses and depth}} \\
\midrule
MegaSaM~\citep{li2025megasam}           & 26.23 & 216.40 & 18.30 & 0.69 & 2.23 \\
ViPE~\citep{huang2025vipe} (default)    & 26.08 & \textbf{215.20} & 18.15 & \textbf{0.72} & 1.92 \\
VGGT-$\Omega$~\citep{wang2026vggtomega} & \textbf{25.93} & 215.87 & \textbf{18.37} & \textbf{0.72} & \textbf{1.88} \\
\bottomrule
\end{tabular}}
\end{table}

\noindent\textbf{Perturbed depth.}
We perturb the per-pixel depth estimated by ViPE~\citep{huang2025vipe} by a controlled percentage. At $5\%$ and $10\%$ the effect is small --- $0.04$ FID and $0.02$ EPE, and $1.06$ FID and $0.18$ EPE respectively --- which spans the error regime of modern depth estimators. Only at $30\%$ does quality degrade appreciably ($5.12$ FID, $0.65$ EPE). This is expected: a perturbation of that magnitude changes the depth ordering of the scene, and therefore which surfaces occlude which, so generation is guided by a geometrically different scene.

\noindent\textbf{Alternative 3D foundation models.}
We also replace the estimated camera poses and depth with those of other state-of-the-art 3D foundation models. UniCaMo works with every estimator tested, and its accuracy tracks that of the estimator: FID and EPE both improve monotonically from MegaSaM~\citep{li2025megasam} through ViPE~\citep{huang2025vipe} to VGGT-$\Omega$~\citep{wang2026vggtomega}, which gives our best overall result ($25.93$ FID, $1.88$ EPE). UniCaMo can therefore inherit future advances in 3D estimation without retraining.

\section{Additional Qualitative Visualizations}
\label{sec:supp_qual}

\newlength{\suppfigwidth}
\setlength{\suppfigwidth}{0.898\linewidth}

\begin{figure}[p]
  \centering
  \includegraphics[width=\suppfigwidth]{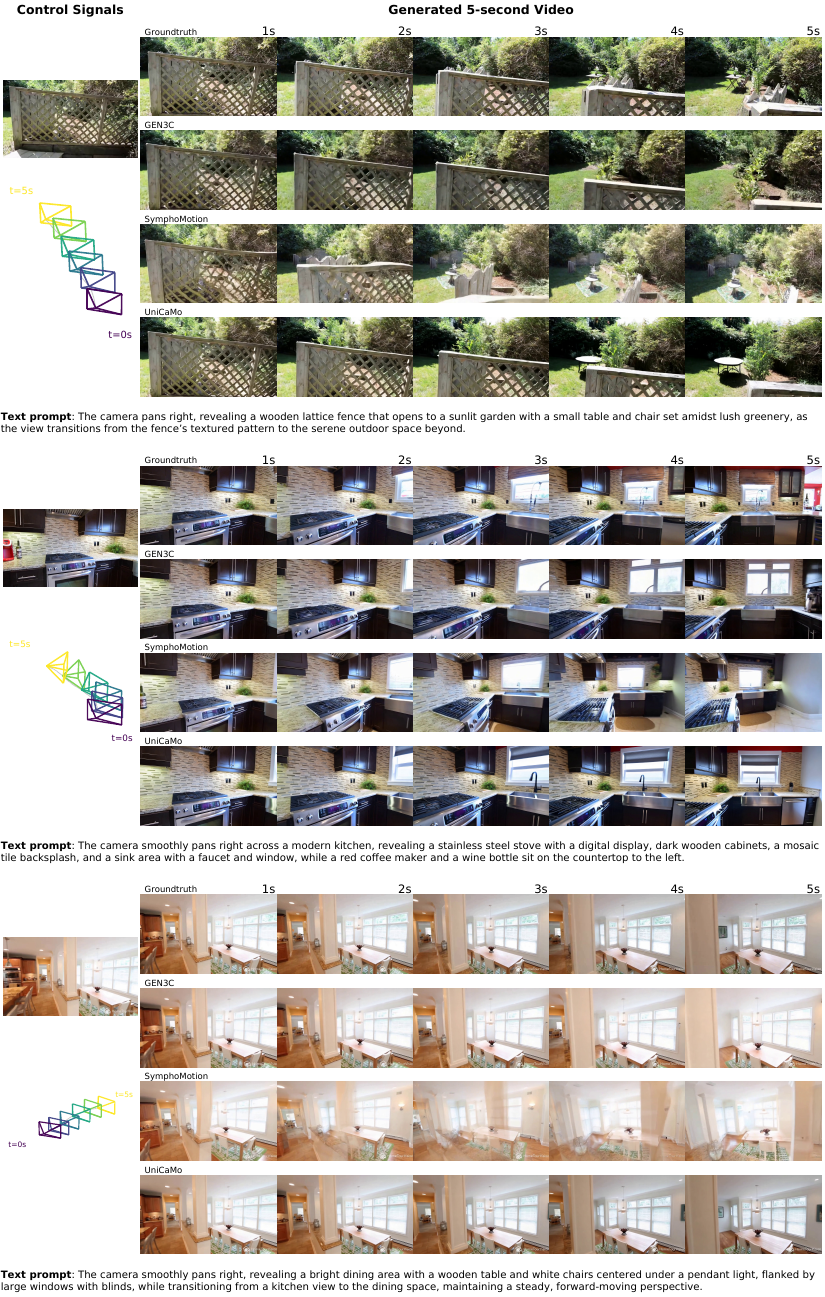}
  \caption{Additional qualitative results for \textbf{camera control} on the
  RealEstate10K dataset.}
  \label{fig:supp_camera_control}
\end{figure}

\begin{figure}[p]
  \centering
  \includegraphics[width=0.95\suppfigwidth]{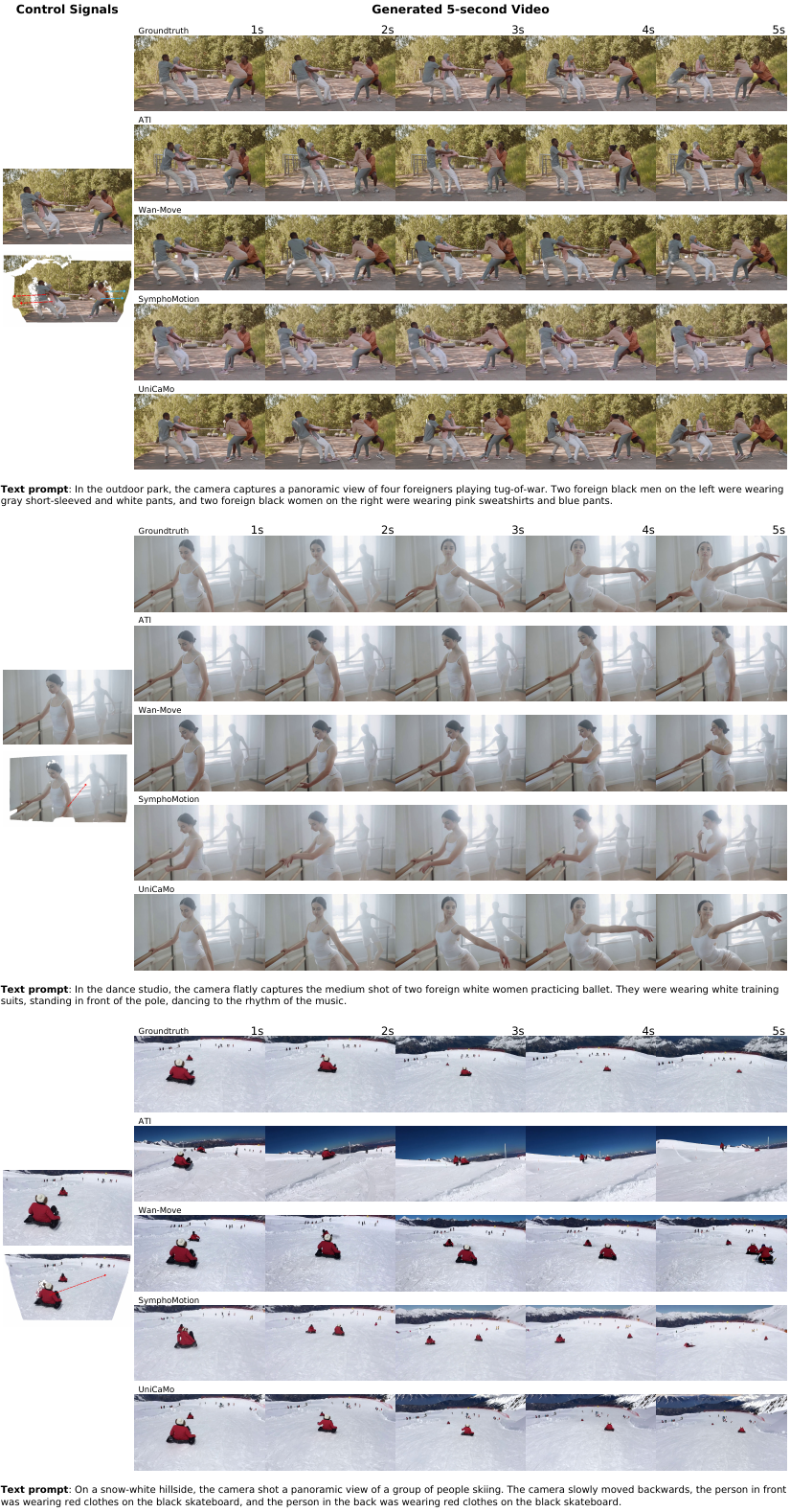}
  \caption{Additional qualitative results for \textbf{motion control} on the
  MoveBench dataset. }
  \label{fig:supp_motion_control}
\end{figure}

\begin{figure}[p]
  \centering
  \includegraphics[width=0.95\suppfigwidth]{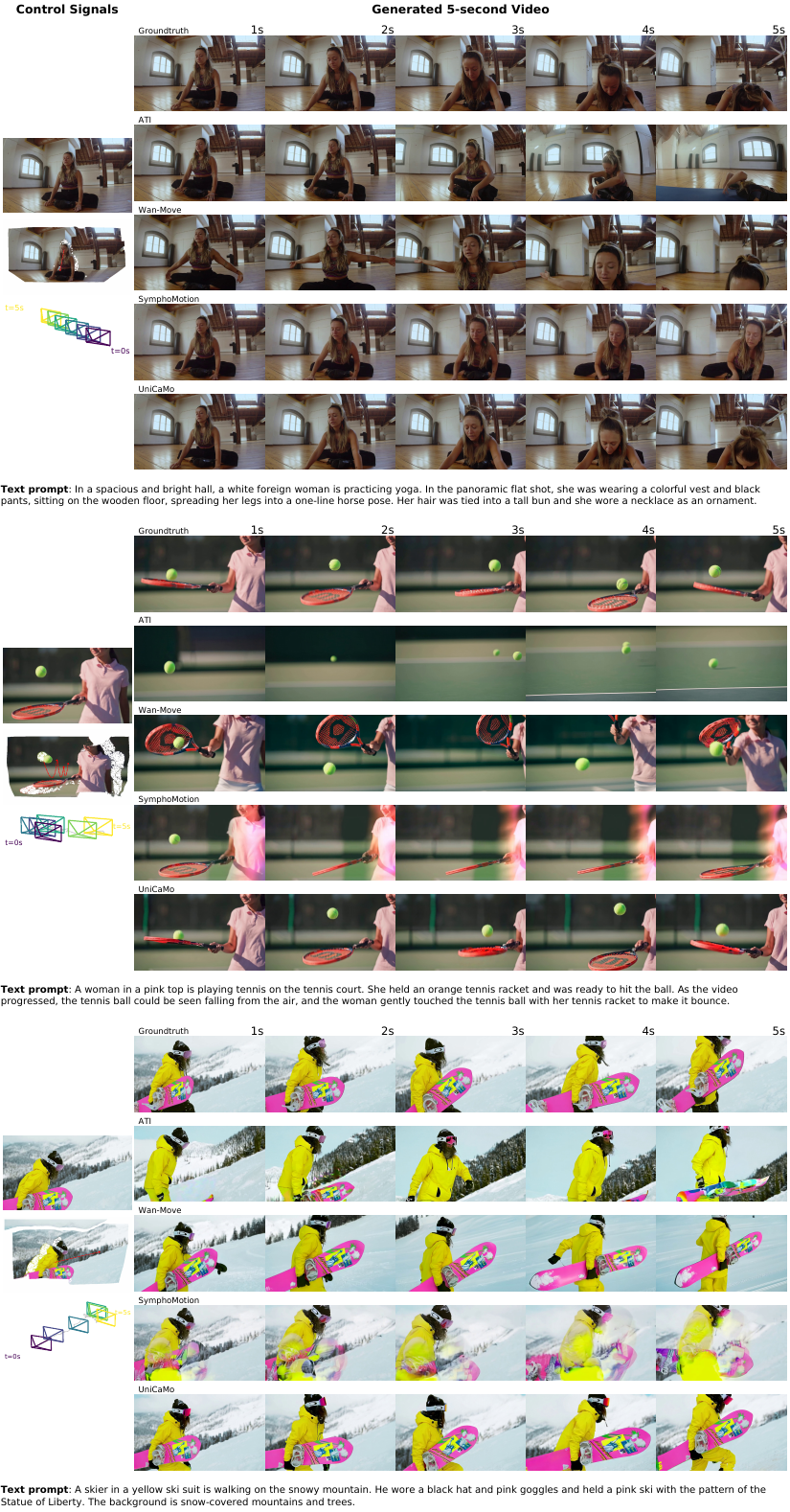}
  \caption{Additional qualitative results for \textbf{joint camera and motion
  control}.}
  \label{fig:supp_join_control}
\end{figure}

\begin{figure}[p]
  \centering
  \includegraphics[width=\suppfigwidth]{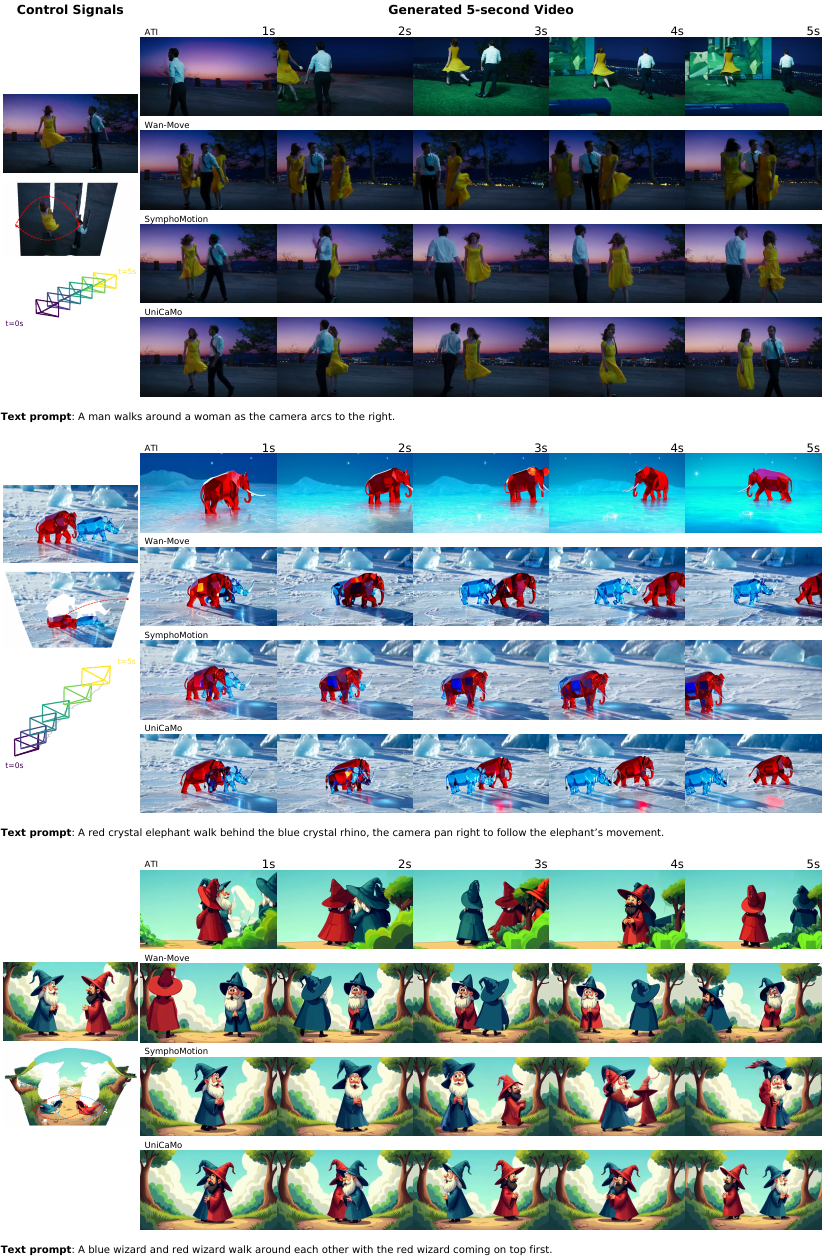}
  \caption{Qualitative results for joint control on \textbf{copmlex trajectory} cases}
  \label{fig:supp_in_the_wild_control}
\end{figure}

In this section, we provide more qualitative results for camera control (see Fig.~\ref{fig:supp_camera_control}), motion control (see Fig.~\ref{fig:supp_motion_control}), joint camera and motion control (see Fig.~\ref{fig:supp_join_control}). Finally, we show our superior results compared with other baseline on some complex trajectory samples in the Fig.~\ref{fig:supp_in_the_wild_control}. Overall, UniCaMo display precise controllable video generation capabilities even in extreme out-of-domain cases.

\clearpage

\section{Discussion, Limitations, and Future Work}
While our 3D-grounded, motion-consistent noise enables flexible control of object and camera motion, the method has limitations. First, the Foreground Noise Propagation stage depends heavily on accurate 3D point tracks and camera poses, so tracking or calibration errors can degrade controllability. Second, the Spherical Noise Sampling stage uses a coarse 3D sphere proxy, which may fail to capture dynamic background geometry. Future improvements in 3D tracking, depth estimation, and noise representations would further enhance our approach.

\end{document}